\pdfoutput=1

\documentclass[11pt]{article}

\usepackage[]{acl2021}

\usepackage{times}
\usepackage{rotating}
\usepackage{adjustbox}
\usepackage{latexsym}
\usepackage{hyperref}
\usepackage{url}
\usepackage{booktabs}
\usepackage{xcolor, colortbl}
\usepackage{graphicx}
\usepackage{multirow}
\usepackage{longtable}
\usepackage[ruled,linesnumbered]{algorithm2e}
\usepackage{amsmath}
\usepackage{algorithmic}

\usepackage[T1]{fontenc}

\usepackage[utf8]{inputenc}

\usepackage{microtype}

\title{Detecting Harmful Online Conversational Content towards LGBTQIA+ Individuals}

\author{Jamell Dacon\Thanks{       Corresponding author}\qquad Harry Shomer\qquad Shaylynn Crum-Dacon\qquad Jiliang Tang\\Michigan State University \\\texttt{\{daconjam, shomerha, crumshay, tangjili\}@msu.edu}}

\begin{document}
\maketitle
\begin{abstract}
{\color{red}{\textit{\textbf{Warning}}}: Due to the overall purpose of the study, this paper contains examples of \textit{stereotypes}, \textit{profanity}, \textit{vulgarity} and other harmful languages in figures and tables that may be triggering or disturbing to LGBTQIA+ individuals, activists and allies, and may be distressing for some readers.}

Online discussions, panels, talk page edits, etc., often contain harmful conversational content \textit{i.e.}, hate speech, death threats and offensive language, especially towards certain demographic groups.
For example, individuals who identify as members of the LGBTQIA+ community and/or BIPOC (Black, Indigenous, People of Color) are at higher risk for abuse and harassment online. In this work, we first introduce a \textit{real-world} dataset that will enable us to study and understand harmful online conversational content. Then, we conduct several exploratory data analysis experiments to gain deeper insights from the dataset. We later describe our approach for detecting harmful online Anti-LGBTQIA+ conversational content, and finally, we implement two baseline machine learning models (i.e., Support Vector Machine and Logistic Regression), and fine-tune 3 pre-trained large language models (BERT, RoBERTa, and HateBERT). Our findings verify that large language models can achieve very promising performance on detecting online Anti-LGBTQIA+ conversational content detection tasks.

\end{abstract}

\section{Introduction}\label{sec:intro}

Harmful online content from \textit{real-word conversations} has become a major issue in today's society, even though queer people often rely on the sanctity of online spaces to escape offline abuse \cite{10.1145/3442442.3452325, DBLP:journals/corr/abs-2109-12192}. However, individuals who may oppose, criticize, or possess contradictory feelings, beliefs, or motivations towards certain communities constitute discrimination, harassment and abuse in the form of hate speech, abusive and offensive language use \cite{10.1145/3442442.3452325, DBLP:journals/corr/abs-2109-12192, blodgett-etal-2020-language, bianchi-hovy-2021-gap}. Unfortunately, this issue results in the maintenance and sustenance of harmful stereotypical societal biases. Online conversational toxicity, death threats and other harmful languages can prevent people from genuinely expressing themselves out of fear of abuse and/or harassment, or encourage self-harm. Conversations pertaining to members of the LGBTQIA+ community may lead to increased feelings of marginalization of an already marginalized community. 

Consequently, social media remains a hostile, exclusive, restrictive, and controlling environment for gender and sexual orientation, race, and LGBTQIA+ individuals, activists and allies \cite{DBLP:journals/corr/abs-2110-09271, 10.1145/3442442.3452325, zmigrod-etal-2019-counterfactual}, despite substantial progress on LGBTQIA+ rights causing a massive surge in negative online behaviors \cite{davidson-etal-2019-racial}. Accordingly, there has been an alarming increase in hate speech and abusive language instances toward the LGBTQIA+ community \cite{DBLP:journals/corr/abs-2110-09271}. \textit{Gender} is a spectrum, where the now LGBTQIA+ acronym continue to expand in an attempt to include all gender and sexual identities, for example, SGL, QPOC/QTPOC, QUILTBAG, etc.\footnote{\url{https://outrightinternational.org/content/acronyms-explained}}. Thereby, there may be as many definitions as there are people. Nonetheless, the LGBTQIA+ community not only considers gender identity, but encapsulates a multiplicity of sexual orientations, and relationships which are integral aspects of our everyday society, yet they lead to discrimination, harassment and abuse both offline and online \cite{DBLP:journals/corr/abs-2109-00227}. 

As conversational platforms struggle to effectively facilitate harmful conversations, there is a dire need to create a safe and inclusive place that welcomes, supports, and values all LGBTQIA+ individuals (with the exception of dating sites and mobile dating applications) \cite{doi:10.1177/0003122412448050} by better facilitating Anti-LGBTQIA+ conversational content \cite{DBLP:journals/corr/abs-2110-09271}. With the expansive growth of artificial intelligence (AI) and natural language processing (NLP) technologies, both researchers and practitioners can perform large-scale analysis, which aids in the automatic interpretation of unstructured text data, \textit{e.g.}, social media comments. To address the above challenges, in this paper, we aim to intersect NLP and queerness by implementing both machine learning (ML) and large language models (LLMs) models to readily identify and detect online Anti-LGBTQIA+ conversational content.\\

\noindent \textbf{Contributions.} The contributions of this work are threefold: 1) we adapt the gender orientation (\textit{LGBTQIA+}, \textit{straight}) dimension from \textsc{RedditBias} \cite{barikeri-etal-2021-redditbias} for the task of harmful conversational content detection to study stereotypical societal biases against LGBTQIA+ individuals by implementing a multi-headed BERT-based toxic comment detection model \cite{Detoxify} to identify several forms of toxicity; (2) a detailed human evaluation of our human annotators to ensure data quality (see Appendix \ref{sec:guide} for details); and (3) we construct a a multi-label classification data for a total of 6 distinct harmful language labels for automated harmful conversational content detection of LGBTQIA+ individuals.

To the best of our knowledge, our dataset is the first such dataset created for both binary and multi-label classification of 6 distinct labels for automated harmful conversational content detection to study stereotypical societal biases against LGBTQIA+ individuals. We intend to release our labeled dataset for future shared tasks of Anti-LGTBQIA+ harmful content detection in hopes that AI and NLP practitioners develop and deploy safe, LGTBQIA+ inclusive technologies to readily identify and remove harmful online conversational content geared toward the LGBTQIA+ community. We release the both the multi-label dataset with all code online at: \url{https://github.com/daconjam/Harmful-LGBTQIA}.

\section{Preliminaries}\label{sec:prelim}

\begin{table*}[t]
\centering
\scalebox{0.9}{\begin{tabular}{lllll}
\toprule
\multicolumn{1}{l}{\textbf{Orientation}} & \multicolumn{4}{l}{\textbf{Comments}} \\ \midrule
\multirow{3}{*}{\begin{tabular}[c]{@{}l@{}}\textit{lesbian}, \textit{gay}, \textit{bisexual},\\ \textit{transgender}, \textit{asexual},\\ \textit{pansexual}, ...\end{tabular}} & \multicolumn{4}{l}{\textit{gay people are faggots and carry a bunch of other diseases with them.}} \\ \cline{2-5} 
 & \multicolumn{4}{l}{\textit{trans lesbians are mentally ill men identifying as women. doesnt sound normal to me}} \\ \cline{2-5} 
 & \multicolumn{4}{l}{\textit{bisexuals are confused. just about life in general}} \\ \bottomrule
\end{tabular}}
\caption{Queerness examples comments from \textsc{RedditBias}.}\label{tab:examples}
\end{table*}

\begin{table*}[t]
\centering
\scalebox{0.9}{\begin{tabular}{l|c|c|c|c|c|c}
\toprule
\textbf{Comments} & \textbf{Toxicity} & \textbf{Severe Toxicity} & \textbf{Obscene} & \textbf{Threat} & \textbf{Insult} & \textbf{Identity Attack} \\ \hline
\multicolumn{1}{l|}{\textit{gay people are faggots...}} & \multicolumn{1}{c|}{0.995} & \multicolumn{1}{c|}{0.214} & \multicolumn{1}{c|}{0.882} & \multicolumn{1}{c|}{0.014} & \multicolumn{1}{c|}{0.953} & 0.777 \\ \hline
\multicolumn{1}{l|}{\textit{trans lesbians are mentally ill...}} & \multicolumn{1}{c|}{0.949} & \multicolumn{1}{c|}{0.042} & \multicolumn{1}{c|}{0.231} & \multicolumn{1}{c|}{0.01} & \multicolumn{1}{c|}{0.446} & 0.711 \\ \hline
\multicolumn{1}{l|}{\textit{bisexuals are confused...}} & \multicolumn{1}{c|}{0.977} & \multicolumn{1}{c|}{0.084} & \multicolumn{1}{c|}{0.271} & \multicolumn{1}{c|}{0.017} & \multicolumn{1}{c|}{0.674} & 0.831 \\ 
\bottomrule
\end{tabular}}
\caption{Automated labeled queerness (shortened) example comments from Table \ref{tab:examples} using Detoxify.}\label{tab:labeled}
\end{table*}

\begin{table*}[t]
\centering
\scalebox{0.9}{\begin{tabular}{c|c|c|c|c|c|c}
\bottomrule
\textbf{Label} & \textbf{Toxicity} & \textbf{Severe Toxicity} & \textbf{Obscene} & \textbf{Threat} & \textbf{Insults} & \textbf{Identity Attack} \\ \hline
1 & 7529 & 185 & 1590 & 28 & 2244 & 4494 \\ \hline
0 & 2401 & 9745 & 8340 & 9902 & 7686 & 5436\\
\bottomrule
\end{tabular}}
\caption{Harmful and non-harmful comment counts w.r.t each label for a total of 9930 comments.}\label{tab:counts}
\end{table*}

In this section, we introduce some preliminary knowledge about the problem under study. We first present the problem statement, then we introduce the dataset and conduct EDA experiments. Later, we describe the automatic labeling process, and human evaluation.

\subsection{Problem Statement}\label{sec:problem}

Due to the rampant use of the internet, there has been a massive surge in negative online behaviors both on social media and online conversational platforms \cite{yin2021towards, DBLP:journals/corr/abs-2109-12192, 10.1145/3442381.3450137, waseem-hovy-2016-hateful}. Hence, there is a great need to drastically reduce hate speech and abusive language instances toward the LGBTQIA+ community to create a safe and inclusive place for all LGBTQIA+ individuals, activists and allies. Therefore, we encourage AI and NLP practitioners to develop and deploy safe and LGTBQIA+ inclusive technologies to identify and remove online Anti-LGBTQIA+ conversational content \textit{i.e.}, if a comment is or contains harmful conversational content conducive to the LGBTQIA+ community \cite{DBLP:journals/corr/abs-2109-00227}.
To address the above problem, we define 3 goals:
\begin{enumerate}
    \item The first goal is to detect several forms of toxicity in comments geared toward LGBTQIA+ individuals such as threats, obscenity, insults, and identity-based attacks.
    
    \item The second goal is to conduct EDA and a detailed human evaluation to gain a better understanding of a new multi-labeled dataset \textit{e.g.}, label correlation and feature distribution that represents the overall distribution of continuous data variables.
    
    \item The third goal is to accurately identify and detect harmful conversational content in social media comments.
\end{enumerate}

\subsection{Dataset}

As Reddit is one of the most widely used online discussion social media platforms, \citet{barikeri-etal-2021-redditbias} released \textsc{RedditBias}, a multi-dimensional societal bias evaluation and mitigation resource for multiple bias dimensions dedicated to conversational AI. \textsc{RedditBias} is created from real-world conversations collected from Reddit, annotated for four societal bias dimensions: (i) Religion (\textit{Jews}, \textit{Christians}) and (\textit{Muslims}, \textit{Christians}), (ii) Race (\textit{African Americans}), (iii) Gender (\textit{Female}, \textit{Male}), and (iv) Queerness (\textit{LGBTQIA+}, \textit{straight}).

We adapt the queerness (\textit{gender/sexual orientation}) dimension and collect a total of 9930 \textit{LGBTQIA+} related comments discussing topics involving individuals who identify as Lesbian, Gay, Bisexual, Transgender, Queer/Questioning, Intersex, Asexual, etc., (see Table~\ref{tab:examples}). 
For more details of \textsc{RedditBias} creation, bias specifications, retrieval of candidates for biased comments, and manual annotation and preprocessing of candidate comments see \cite{barikeri-etal-2021-redditbias}. In addition, \textsc{RedditBias} is publicly available with all code online at: \url{https://github.com/umanlp/RedditBias}.

\subsection{Annotation} \label{sec:annotation}

Our collected dataset encapsulates a multitude of gender identities and sexual orientations, thus, we denote our dataset under the notion of \textit{queerness}. Note that the comments from \textsc{RedditBias} are unlabeled for our tasks, hence we attempt to label each comment accordingly for each classification task. To do so, we implement Detoxify\footnote{\url{https://github.com/unitaryai/detoxify}} \cite{Detoxify}, a multi-headed BERT-based model \cite{DBLP:journals/corr/abs-1810-04805} capable of detecting different types of toxicity such as \textit{threats}, \textit{obscenity}, \textit{insults}, and \textit{identity-based attacks} and discovering unintended bias in both English and multilingual toxic comments. Detoxify is created using Pytorch Lightning\footnote{https://www.pytorchlightning.ai} and Transformers\footnote{https://huggingface.co/docs/transformers/index}, and fine-tuned on datasets from 3 Jigsaw challenges, namely Toxic comment classification, Unintended Bias in Toxic comments and Multilingual toxic comment classification for multi-label classification to detect toxicity across a diverse range of conversations. 

Specifically, we use Detoxify's $\mathrm{original}$ model that is trained on a 
large open dataset of Wikipedia+Civil Talk Page comments which have been labeled by human raters for toxic behavior. 
In Table~\ref{tab:labeled}, we display \textit{harmful} predicted probabilities of (shortened) example comments from Table \ref{tab:examples} into their respective labels (\textit{i.e.,} $\mathsf{toxicity}$, $\mathsf{severe}$ $\mathsf{toxicity}$, $\mathsf{obscene}$, $\mathsf{threat}$, $\mathsf{insult}$, and $\mathsf{identity}$ $\mathsf{attack}$), using Detoxify. Therefore, we create a multi-labeled queerness dataset which can be used for downstream tasks such as binary and multi-label toxic comment classification to predict Anti-LGBTQIA+ online conversational content.

\begin{figure}[!htbp]
    \centering
    \includegraphics[scale=0.43]{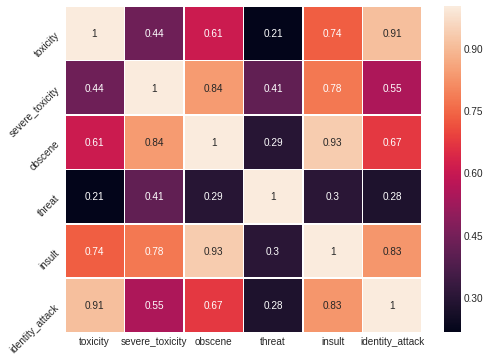}
    \caption{Label correlation heatmap matrix.}
    \label{fig:correlation}
\end{figure} 
Unfortunately, to avoid public discrepancies such as cultural and societal prejudices amongst human raters, competitors, and LGBTQIA+ individuals of the 3 Jigsaw challenges and its data, official documentation and definitions of these classifications are unavailable. Therefore, we cannot know for certain what each label (\textit{l}) means and why. Therefore, to address the issue of this ``unknown'' labeling schema and gain better datasets insights, such as feature distributions and identifying what quantifies a comment as ``harmful'' or ``non-harmful'', 
we conduct both univariate and multivariate analysis. In Figure~\ref{fig:correlation}, we illustrate a correlation matrix between labels.
For each cell, we follow a label threshold setting \textit{i.e.,} $l >$ 0.7 to determine heavily correlated labels.
Here, we see several heavily correlated relationships (i) $\mathsf{toxicity}$, $\mathsf{insult}$ and $\mathsf{identity}$ $\mathsf{attack}$, and (ii) $\mathsf{severe}$ $\mathsf{toxicity}$, $\mathsf{insult}$ and $\mathsf{obscene}$.

As previously mentioned, our automated labeled queerness dataset contains probabilities of comments' harmfulness. For each classification task, we will now consider two classes, ``harmful : 1'' and ``non-harmful : 0'' per label by following a class (\textit{c}) threshold mapping system \textit{i.e.,} $c >=$ 0.5 $\rightarrow$ 1 to determine whether a comment is deemed harmful, or not. In Table~\ref{tab:counts}, we display harmful comment counts w.r.t each label that satisfy our class threshold mapping system. 

More information about figures on data breakdowns, distribution plots (\textit{i.e.}, to depict the variation in the data distribution), and knowledge of which words constitute a ``harmful'' or ``non-harmful'' comment for each label can be found in 
Appendices~\ref{sec:breakdown}, \ref{sec:fd} and \ref{sec:word}.

\subsection{Human Evaluation}

We employ Amazon Mechanical Turk (AMT) annotators\footnote{Each AMT annotators is independent, and either an LGBTQIA+ individual, activist or ally. In addition, each annotator is filtered by HIT approval rate $\geq$ 93\%, completed $>$ 7,500 HITs and located within the United States.}. Due to the nature of the comments, it was quite difficult to acquire a large number of annotators that were willing to manually rate 1000 randomly sampled comments to measure the effectiveness of the toxicity classifier, due to the examples of \textit{stereotypes}, \textit{profanity}, \textit{vulgarity} and other harmful languages towards LGBTQIA+ individuals. Note that terms are not filtered as they are representative of real-world conversations and are exceedingly essential to our goals mentioned in Section~\ref{sec:problem}. After this, a total 15 annotators (maximum) were more than willing to help achieve this goal. 

First, we aggregate an LGBTQIA+ sources from OutRight\footnote{\url{https://outrightinternational.org}}, a human rights organization for LGBTQIA+ people in an attempt to educate annotators on identity, sexuality, and relationship definitions of the expanded LGBT (older) acronym in a move towards inclusivity. Then, the annotators were asked to indicate whether a comment is \textit{toxic} or \textit{non-toxic}. As the $\mathsf{toxicity}$ label is the most prevalent label, if a comment $x$ is deemed \textit{non-toxic}, then the annotators may discard this comment. However, if $x$ is deemed \textit{toxic}, then each annotator is provided with 5 additional labels (\textit{i.e.,} $\mathsf{severe}$ $\mathsf{toxicity}$, $\mathsf{obscene}$, $\mathsf{threat}$, $\mathsf{insult}$, and $\mathsf{identity}$ $\mathsf{attack}$) along with their respective definitions\footnote{Each definition is aggregated from a trusted online dictionary -- \url{https://www.dictionary.com}} to determine if a comment is or contains Anti-LGBTQIA+ content. 

To ensure the label quality of our samples and to quantify the extent of agreement between raters, we first let each rater independently rate each comment considering two classes, ``harmful : 1'' and ``non-harmful : 0'', then we measure the inter-annotator agreement (IAA) using Krippendorff's $\alpha$. Then, we calculate a Krippendorf's $\alpha$ of 0.81 using NLTK’s \cite{loper-bird-2002-nltk} $\mathrm{nltk.metrics.agreement}$, and did not observe significant differences in agreement across the individual comments and the implementation of Detoxify. Annotators informed us of particular $\mathsf{severe}$ $\mathsf{toxicity}$ cases that may arise if an annotator identifies as a member of the LGBTQIA+ community is triggered by their personal experiences regarding in-person and online discrimination or abuse based on their actual or perceived sexual orientation, gender identity or expression. The full annotation guidelines can be found in Appendix \ref{sec:guide}.

\section{Experimental Settings}\label{sec:exp_settings}
In this section, we will introduce both our proposed out-of-the-box ML models \textit{i.e.}, Logistic Regression and Support Vector Machine, and LLMs \textit{i.e.}, BERT, RoBERTa and HateBERT. Next, we outline our two classification tasks, namely, binary and multi-label classification and detail each model accordingly.

\subsection{Baseline Models}

\begin{itemize}
    \item \textbf{Logistic Regression} - Logistic Regression (LGR) \cite{wright1995logistic, kleinbaum2002logistic, menard2002applied} is a statistical ML model that exploits linear combinations of one or more independent variables to predict an output. In regression analysis, LGR is a memory-based predictive tool that considers the probability of one event occurring, thus estimating the parameters of a logistic model to determine when a dependent variable is dichotomous (binary classification). 
    
    \item \textbf{Support Vector Machine} - Support Vector Machine (SVM) \cite{noble2006support} is one of the most powerful out-of-the-box supervised ML algorithms. Given a dataset $\mathcal{D}$, SVM attempts to compute $y = |w \cdot x + b|$ for the smallest $y$ to discover boundaries between classes, and is unaffected by outliers.  
\end{itemize}

First, we use a Natural Language Toolkit (NLTK) \cite{loper-bird-2002-nltk} to remove stop words, then we implement two well-known linear scikit-learn\footnote{\url{https://scikit-learn.org/stable/modules/multiclass.html}} models with parameters \textit{solver=`sag'} for our LGR model\footnote{Stochastic Average Gradient: SAG method optimizes the sum of a finite number of smooth convex functions by incorporating previous gradient values to achieve a faster convergence rate than black-box SG methods.}, and default kernel \textit{C=1.0} for our SVM model. Next, we utilize a feature extraction method, namely term-frequency inverse document frequency\footnote{A commonly used numerical statistic technique that generates a numerical weighting of words which signifies the importance of each word in a given corpus document.} (TF-IDF) and implement a $\mathsf{One-Vs-Rest}$ multi-class classification heuristic method to divide our multi-class classification task into multiple binary classification datasets as our baseline models are designed for binary classification problems.

\subsection{Deep Learning Models}

We use 3 different LLMs for our tasks. We select 3 popular LLMs (BERT \cite{DBLP:journals/corr/abs-1810-04805}, RoBERTa \cite{DBLP:journals/corr/abs-1907-11692} and HateBERT \cite{DBLP:journals/corr/abs-2010-12472}. Specifically, HateBERT \cite{DBLP:journals/corr/abs-2010-12472} has demonstrated superior hate speech and abusive detection \cite{kalyan2021ammus, yin2021towards}. We will now detail our selected LLMs below:

\begin{itemize}
    \item \textbf{BERT} - BERT \cite{DBLP:journals/corr/abs-1810-04805} is a bidirectional transformer \cite{transformer} encoder used to learn language representations for a variety of downstream tasks by pre-training via two tasks: (i) masked language modeling (MLM) and (ii) next sentence prediction (NSP). This allows BERT to learn a general set of weights to model a variety of different inputs. BERT can then be fine-tuned on new datasets and tasks by training an additional layer on top of BERT.  
    
    \item \textbf{RoBERTa} - RoBERTa \cite{DBLP:journals/corr/abs-1907-11692} attempts to improve upon BERT via several modifications. This includes removing NSP pre-training task, training with more data, and modifying the MLM pre-training task. \citet{DBLP:journals/corr/abs-1907-11692} modify the MLM task by considering dynamic masking. This, as opposed to BERT, only uses a single static mask for each input.
    
    \item \textbf{HateBERT} - HateBERT \cite{DBLP:journals/corr/abs-2010-12472} attempts to fine-tune BERT to detect abusive and hateful language. \citet{DBLP:journals/corr/abs-2010-12472} do so by first constructing the RAL-E (Reddit Abusive Language English) dataset, which contains Reddit posts. The authors only consider posts from subreddits that were banned for hateful or offensive language. \citet{DBLP:journals/corr/abs-2010-12472} retrain BERT on RAL-E using its original weights resulting in a more robust model attuned to understanding hateful speech. 
    
\end{itemize}

Each of the three LLMs is fine-tuned for 5 epochs with a learning rate of $5e^{-5}$, a batch size of 8, and optimized using AdamW \cite{adamw}. Each model has a maximum sequence length of 512 tokens, therefore, longer sequences are truncated while shorter sequences are padded.

\begin{table*}[t]
\centering
\scalebox{0.95}{\begin{tabular}{c|c|c|c|c}
\toprule
\multirow{2}{*}{\textbf{Model}} & \multicolumn{4}{c}{\textbf{Binary Classification}} \\ \cline{2-5}
 & \textbf{Class} & \textbf{Precision} & \textbf{Recall} & \textbf{F1-Score} \\ \hline
\multirow{2}{*}{LGR} & 1 & 0.85 &  \textbf{0.97} & 0.9 \\ 
 & 0 & 0.82 & 0.49 & 0.61  \\ \hline
\multirow{2}{*}{SVM} & 1  & 0.88 &  0.93 & 0.9   \\ 
 & 0  & 0.74  & 0.61  & 0.67 \\ \hline
\multirow{2}{*}{BERT} & 1 &  \textbf{0.94} & 0.96 & \textbf{0.95} \\ 
 & 0 &  0.86 & \textbf{0.83}  & \textbf{0.84} \\ \hline
\multirow{2}{*}{RoBERTa} & 1 &  0.92 &  0.96 &  0.94 \\  
 & 0 & 0.86 & 0.76 & 0.81 \\ \hline
\multirow{2}{*}{HateBERT} & 1 &  0.93 & 0.96 & 0.94 \\  
 & 0 &  \textbf{0.87} &  0.77 & 0.82 \\
\bottomrule
\end{tabular}}
\caption{Classification report \textit{only} on Toxicity label for the classes Toxic (1) and Non-toxic (0).}\label{tab:toxic_binary}
\end{table*}

\subsection{Classification Tasks}

\subsubsection{Binary  Classification}

The goal of our baseline models is to implement traditional ML algorithms to assess the difficulty of the simpler of our two classification tasks, and provide a standard to which our deep learning models can be compared. We aim to conduct a straightforward \textit{binary} classification task by isolating $\mathsf{toxicity}$ labeled comments. Given a comment $x$, a binary output is assigned to each class $\forall l$ \textit{i.e.}, \textit{toxic} class is indicated with a 1 and \textit{non-toxic} class with a 0. 

\subsubsection{Multi-label Classification}
Our \textit{multi-label} classification task is to classify each $x$ with $n$ labels (\textit{i.e.,} $\mathsf{toxicity}$ (T), $\mathsf{severe}$ $\mathsf{toxicity}$ (S), $\mathsf{obscene}$ (O), $\mathsf{threat}$ (Th), $\mathsf{insult}$ (In), and $\mathsf{identity}$ $\mathsf{attack}$ (Id)), that are \textbf{not mutually exclusive} (see Figure \ref{fig:correlation}). Each $x$ is assigned to each class $\forall l$ i.e., \textit{harmful} class is indicated with a 1 and \textit{non-harmful} class with a 0.

\subsection{Evaluation Metrics}

To evaluate the performance of each model on the test set, we report the precision, recall and F1 score. The precision is the percentage of samples classified as positive that are truly positive, while the recall is defined as the true positive rate. The F1 score is the harmonic mean of precision and recall.

\section{Experimental Results}\label{sec:results}

In this section, we conduct experiments to verify the effectiveness of the different models for both classification tasks following the experimental settings introduced in Section \ref{sec:exp_settings}. For each task, we present the performance of predicting each type of label. Furthermore, for the binary classification task, we further detail the overall performance.

\subsection{Binary Classification}

In this subsection, we evaluate both baseline and deep learning models on the binary classification task. This task aims to predict the toxicity of a comment. The performance metrics for each class and overall performance metrics on the test set are shown in Table \ref{tab:toxic_binary} and  Table \ref{tab:toxic_f1}, respectively. 

\begin{table}[!htbp]
\centering
\scalebox{0.95}{\begin{tabular}{c|c|c}
\toprule
\multirow{3}{*}{\textbf{Model}} & \multicolumn{2}{c}{\textbf{Binary Classification}} \\ \cline{2-3}
 & \multicolumn{1}{c|}{\begin{tabular}[c]{@{}c@{}}\textbf{macro}\\ \textbf{F1-score}\end{tabular}} & \begin{tabular}[c]{@{}c@{}}\textbf{weighted}\\ \textbf{F1-score}\end{tabular} \\ \hline
{LGR} & 0.76  &  0.83 \\ \hline
{SVM} & 0.79  &  0.84   \\ \hline
{BERT} & \textbf{0.9}   & \textbf{0.92} \\ \hline
{RoBERTa} &  0.87 & 0.91  \\\hline
{HateBERT} &  0.88 & 0.91 \\
\bottomrule
\end{tabular}}
\caption{Macro and weighted F1-score \textit{only} on Toxicity label.}\label{tab:toxic_f1}
\end{table}

\begin{table*}[t]
\centering
\scalebox{0.9}{\begin{tabular}{cc|ccccc|c|c}
\toprule
 & \multicolumn{8}{c}{\textbf{Multi-label Classification}} \\ \hline
\multirow{2}{*}{\textbf{Label}} & \multirow{2}{*}{\textbf{Class}} & \multicolumn{5}{c|}{\textbf{Models}} & \multirow{2}{*}{\textbf{Train}} & \multirow{2}{*}{\textbf{Test}} \\\cline{3-7}
&  & \multicolumn{1}{c|}{\textbf{LGR}} & \multicolumn{1}{c|}{\textbf{SVM}} & \multicolumn{1}{c|}{\textbf{BERT}} & \multicolumn{1}{c|}{\textbf{RoBERTa}} & \textbf{HateBERT} &  & \\ \hline

\multirow{2}{*}{T} & \multicolumn{1}{c|}{1} & \multicolumn{1}{c|}{0.88} & \multicolumn{1}{c|}{0.68} & \multicolumn{1}{c|}{\textbf{0.96}} & \multicolumn{1}{c|}{0.95} & \multicolumn{1}{c|}{0.95} & \multicolumn{1}{c|}{5969} & 1490 \\ 

 & \multicolumn{1}{c|}{0} & \multicolumn{1}{c|}{0.49} & \multicolumn{1}{c|}{0.56} & \multicolumn{1}{c|}{\textbf{0.87}} & \multicolumn{1}{c|}{0.84} & \multicolumn{1}{c|}{0.84} & \multicolumn{1}{c|}{1975}  & 496 \\  
 \hline
 \multirow{2}{*}{S} & \multicolumn{1}{c|}{1} & \multicolumn{1}{c|}{0.35} & \multicolumn{1}{c|}{0.08} & \multicolumn{1}{c|}{\textbf{0.82}} & \multicolumn{1}{c|}{0.71} & \multicolumn{1}{c|}{0.72} & \multicolumn{1}{c|}{148} & 37 \\
 
 & \multicolumn{1}{c|}{0} & \multicolumn{1}{c|}{0.99} & \multicolumn{1}{c|}{0.75} & \multicolumn{1}{c|}{\textbf{1}} & \multicolumn{1}{c|}{0.99} & \multicolumn{1}{c|}{\textbf{1}} & \multicolumn{1}{c|}{7796} & 1949\\ 
 
 \hline
 \multirow{2}{*}{O} & \multicolumn{1}{c|}{1} & \multicolumn{1}{c|}{0.7} & \multicolumn{1}{c|}{0.44} & \multicolumn{1}{c|}{\textbf{0.91}} & \multicolumn{1}{c|}{\textbf{0.91}} & \multicolumn{1}{c|}{\textbf{0.91}} & \multicolumn{1}{c|}{1295} & 295 \\
 
 &  \multicolumn{1}{c|}{0} & \multicolumn{1}{c|}{0.96} & \multicolumn{1}{c|}{0.78} & \multicolumn{1}{c|}{\textbf{0.98}} & \multicolumn{1}{c|}{\textbf{0.98}} & \multicolumn{1}{c|}{\textbf{0.98}} & \multicolumn{1}{c|}{6649} & 1691\\ 
 
 \hline
 \multirow{2}{*}{Th} & \multicolumn{1}{c|}{1} & \multicolumn{1}{c|}{0} & \multicolumn{1}{c|}{\textbf{0.01}} & \multicolumn{1}{c|}{0} & \multicolumn{1}{c|}{0} & \multicolumn{1}{c|}{0} & \multicolumn{1}{c|}{22} & 6 \\ 
 
 & \multicolumn{1}{c|}{0} & \multicolumn{1}{c|}{0.74} & \multicolumn{1}{c|}{\textbf{1}} & \multicolumn{1}{c|}{\textbf{1}} & \multicolumn{1}{c|}{\textbf{1}} & \multicolumn{1}{c|}{\textbf{1}} & \multicolumn{1}{c|}{7722} & 1980\\ 
 
 \hline 
 \multirow{2}{*}{In} & \multicolumn{1}{c|}{1} & \multicolumn{1}{c|}{0.64} & \multicolumn{1}{c|}{0.59} & \multicolumn{1}{c|}{\textbf{0.89}} & \multicolumn{1}{c|}{0.87} & \multicolumn{1}{c|}{0.87} & \multicolumn{1}{c|}{1804} & 440  \\
 
 & \multicolumn{1}{c|}{0} & \multicolumn{1}{c|}{0.93} & \multicolumn{1}{c|}{0.81} & \multicolumn{1}{c|}{\textbf{0.97}} & \multicolumn{1}{c|}{0.96} & \multicolumn{1}{c|}{0.96} & \multicolumn{1}{c|}{6140} & 1546\\ 
 
 \hline 
 \multirow{2}{*}{Id} & \multicolumn{1}{c|}{1} & \multicolumn{1}{c|}{0.76} & \multicolumn{1}{c|}{0.76} & \multicolumn{1}{c|}{\textbf{0.89}} & \multicolumn{1}{c|}{\textbf{0.89}} & \multicolumn{1}{c|}{\textbf{0.89}} & \multicolumn{1}{c|}{3599} & 895  \\
 
 & \multicolumn{1}{c|}{0} & \multicolumn{1}{c|}{0.82} & \multicolumn{1}{c|}{0.82} & \multicolumn{1}{c|}{\textbf{0.92}} & \multicolumn{1}{c|}{0.91} & \multicolumn{1}{c|}{0.91} & \multicolumn{1}{c|}{4345} & 1091 \\
\bottomrule
\end{tabular}}
\caption{F1-score results on \textit{all} labels for the classes Harmful (1) and Non-harmful (0) into their respective labels (\textit{i.e.,} $\mathsf{toxicity}$ (T), $\mathsf{severe}$ $\mathsf{toxicity}$ (S), $\mathsf{obscene}$ (O), $\mathsf{threat}$ (Th), $\mathsf{insult}$ (In), and $\mathsf{identity}$ $\mathsf{attack}$ (Id)), and their statistics.} \label{tab:multi_exps}
\end{table*}

We first investigate the performance in predicting each specific class. The results are detailed in Table \ref{tab:toxic_binary}. Our proposed BERT model achieves the best overall performance for both classes in F1 score. It also achieves the best performance when predicting the toxic comments (i.e. class 1) in precision and predicting non-toxic comments (i.e. class 0) in recall \textit{i.e.}, BERT attains the best performance when predicting class 1 versus class 0 compared to each model. This is likely due to the imbalanced nature of the dataset. As shown in Table \ref{tab:counts}, the number of samples of class 1 outnumber those belonging to class 0 by about 3 to 1. This imbalance may help the models learn to better categorize toxic comments as opposed to non-toxic comments.

We further evaluate the overall performance shown in Table \ref{tab:toxic_f1}. We measure the overall performance using both the macro and weighted F1-score. There is a clear distinction between the performance of the deep learning and baseline models. The performance of the baseline models is similar and lies in the range of 0.76-0.79 and 0.83-0.84 for the macro and weighted F1 scores, respectively. On the other hand, the deep learning models produce higher model performance in the range of 0.87-0.9 and 0.91-0.92 for the macro and weighted F1 scores, respectively. Specifically, we see that the best performing deep learning model, BERT, performs $ \sim 10$\% higher in both metrics versus the best performing baseline model SVM.

\subsection{Multi-label Classification}

In this subsection, we evaluate both baseline and deep learning models for the multi-label classification task. This task aims to predict 6 distinct harmful comment labels. The performance metrics for each class and overall performance metrics on the test set are shown in Table \ref{tab:multi_exps}, respectively.

Previously mentioned in Section \ref{sec:exp_settings}, we first train and fine-tune 5 ML and DL models for our multi-label classification task. In Table \ref{tab:multi_exps}, we display the F1 score for each label considering a two class system such as harmful (class 1) and non-harmful (class 0). Although, the ML models took little to no time to train compared to the LLMs, the DL models outperform the baseline models on the test set. Similarly to the binary classification task, BERT achieves the best performance. 

In particular, BERT attained the best F1 score in all but one instance as there exists a large disparity between the performance for the $\mathsf{threat}$ for both classes. For this label, most of the models achieve perfect performance for class 0, while for class 1 only the SVM model achieves an F1-score above 0. This is most likely caused by the massive class imbalance present for that label, as only .28\% of comments are considered threatening causing the models to classify all comments as non-threatening, thereby resulting in disparate performance.

\section{Discussion}\label{sec:discuss}

In the previous section, we detail the modle performance results for our two classification tasks. Our results indicate that the BERT model fine-tuned on the \textit{queerness} dataset for each task performs best. Now, the question arises, ``\textit{Why does BERT outperform HateBERT, a model designed specifically for detecting hate speech?} We assume that it could be due to the difference in tasks. Specifically, the RAL-E dataset \cite{DBLP:journals/corr/abs-2010-12472}, in which HateBERT is retrained on instead of the 3.3 billion word Books Corpus + Wikipedia dataset BERT \cite{DBLP:journals/corr/abs-2010-12472} is trained on, is created by collecting posts from subreddits banned for several types offensive or abusive language. However, we create our dataset from only those comments REDDITBIAS dataset \cite{barikeri-etal-2021-redditbias} deemed biased against LGBTQIA+ individuals. It is possible that the RAL-E dataset contains few examples of harmful language directed to LGBTQIA+ individuals, and thus, making the resulting model not as useful to our tasks. For such reason, training the original BERT allows a more flexible set of initial parameters to adapt for our experiments. 

Furthermore, we find results that may be of interest among several labels for our multi-label classification
As seen in Table \ref{tab:multi_exps}, labels such as $\mathsf{toxicity}$ (T), $\mathsf{obscene}$ (O), $\mathsf{insult}$ (In), and $\mathsf{identity}$ $\mathsf{attack}$ (Id) achieve an F1-Score of at least 0.87 when detecting both harmful and non-harmful comments. However, when attempting to predict harmful comments that are labeled $\mathsf{severe}$ $\mathsf{ toxic}$ (S) and $\mathsf{threat}$ (Th), whether harmful or not the models performs worse as a result of a class imbalance. For example, only about 2\% of training samples are considered harmful for the $\mathsf{severe}$ $\mathsf{toxic}$ (S) label. Such large imbalances skew the classifiers to more likely predict comments as being non-harmful, causing lower performance on the test set. Future work can consider imbalance strategies such as oversampling or sample weighting to alleviate such issues. 
\section{Related Work}\label{sec:related}

Several works have been proposed for the task of harmful content detection such as gender bias, toxicity, offensiveness, abusiveness, and hate speech on social media (see \citet{fortuna-etal-2020-toxic} for definitions). Starting with one of the most popular social media platforms, Twitter, there has been numerous works of hate speech and toxic language detection \cite{Founta_2018, nejadgholi-cross, waseem-etal-2017-understanding, fortuna-etal-2020-toxic, 10.1145/3442188.3445875, razo-kubler-2020-investigating, zampieri-etal-2019-semeval, waseem-hovy-2016-hateful}. In addition, several works have used data from other social media platforms, such as Reddit \cite{razo-kubler-2020-investigating, wang-etal-2020-detect, zampieri-etal-2019-semeval}, Facebook \cite{glavas-etal-2020-xhate, zampieri-etal-2019-semeval}, Whatsapp \cite{10.1145/3442381.3450137}. 

Works have also attempt to tackle this problem exploiting data from other online conversational platforms such as news media \cite{10.1145/3442442.3452325, razo-kubler-2020-investigating, zampieri-etal-2019-semeval}, Wikipedia discussions/talk page edits and bios \cite{10.1145/3485447.3512134, fortuna-etal-2020-toxic, glavas-etal-2020-xhate, nejadgholi-cross}, and online discussion forums and chatrooms \cite{wang-etal-2020-detect, gao-etal-2020-offensive}.

Over the past years, there has been an increased focus on addressing societal bias issues related to language around gender \cite{zmigrod-etal-2019-counterfactual, 10.1145/3442442.3452325, rudinger-etal-2018-gender, DBLP:journals/corr/abs-1910-10486} and racial bias \cite{davidson-etal-2019-racial, DBLP:journals/corr/BlodgettO17, DBLP:journals/corr/abs-2005-12246, DBLP:journals/corr/abs-1910-10486}. Meanwhile, there has been relatively little work on societal bias such as toxicity, offensive and abusive language, and hate speech detection for LGBTQIA+ individuals \cite{DBLP:journals/corr/abs-2109-00227}. Recently, NLP researchers are narrowing their focused work to detect and measure online harmfulness against LGBTQIA+ individuals \cite{barikeri-etal-2021-redditbias, DBLP:journals/corr/abs-2109-00227}. Specifically, \citet{DBLP:journals/corr/abs-2109-00227} provides a new hierarchical taxonomy for online homophobia and transphobia detection, as well as an homophobic/transphobic labeled dataset which comprises 15,141 annotated multilingual Youtube comments.

\section{Conclusion}\label{sec:conclusion}

In this study, we conducted a systemic evaluation of harmful comments geared towards the LGBTQIA+ community by real Reddit users by isolating and labeling the queerness (\textit{LGBTQIA+}, \textit{straight}) dimension from \textsc{RedditBias} \cite{barikeri-etal-2021-redditbias}. We exploit two out-of-the-box ML methods and 3 state-of-the-art large language models to evaluate the harmfulness of comments and our findings show promising results. 

Despite HateBERT \cite{DBLP:journals/corr/abs-2010-12472} being pre-trained for hate speech and abusive language detection, BERT outperformed all our proposed baseline models and LLMs achieving an F1-score $\geq$ 0.82 for each label, with the exception of the $\mathsf{threat}$ (Th) label due to the extremely low number of \textit{harmful} samples in our multi-label classification task. While this dataset may be useful to understand societal biases, we must warrant mentioning that examples containing \textit{stereotypes}, \textit{profanity}, \textit{vulgarity} and other harmful language that were categorized into 6 distinct classes may be more offensive, triggering or disturbing than the numbers reflected in both the paper and dataset to LGBTQIA+ individuals.

We intend to release our multi-labeled dataset for future shared tasks of online Anti-LGTBQIA+ content detection. We hope that our contributions can constitute potential groundwork on ways that LGBTQIA+ relevant data can effectively be integrated in conversational platforms (e.g., Reddit, Twitter, Facebook) to better facilitate harmful online conversational content. In the future, we aim to explore bias mitigation solutions towards members of the LGBTQIA+ community with the objective of creating a safe and inclusive place that welcomes, supports, and values all LGBTQIA+ individuals, activists and allies both online and offline.

\section{Limitations and Ethical Considerations}\label{ethics}

Our work solely calls for NLP practitioners to prioritize collecting a set of training data with several examples to detect harmful online content by interpreting visual content in context. All authors declare that all data used is publicly available and do not intend for increased feelings of marginalization. We solely attempt to highlight the need for impactful speech and language technologies to readily identify and quickly remove harmful online content to minimize further stigmatization of an already marginalized community. 

Words associated with swearing, insults, or profanity that are present in a comment such as ``\textit{lol bro you gay as fuck lmao!}'' will be classified as toxic, regardless of the tone or the intent of the author \textit{e.g.}, humorous/self-deprecating conversations. Thus, our approach may present some biases toward already vulnerable minority \textit{e.g}, L1 speakers of African American English (AAE) such as BIPOC (Black, Interracial, or People of Color) individuals \cite{dacon2022towards} which may further censor LGBTQIA+ people online.
Therefore, combatting LGBTQIA+ online abuse is particularly challenging, where general-purpose content moderation can be harmful. 

All authors understand that findings in this study may consequently lead to other ethical issues as (online) AAE can incorporate both slang and internet-influenced terms which contain several informal phrases, expressions, idioms, and cultural and regional-specific lingo \cite{blodgett-etal-2020-language, dacon2022towards}. Lastly, we must mention that the opinions that are presented in this paper solely represent the views of the authors, and do not necessarily reflect the official policies of their respective organizations.

\section{Acknowledgements}\label{acks}

This work was made possible in part by the support of both Jamell Dacon and Harry Shomer from National Science Foundation DGE-1828149. The authors would like to thank the anonymous reviewers for their constructive comments.

\bibliography{main}
\bibliographystyle{acl_natbib}

\appendix

\section{Annotation Guidelines}\label{sec:guide}

First, you will be given an extensive list of acronyms and terms from OutRight (Link: \url{https://outrightinternational.org/content/acronyms-explained}), an LGBTQIA+ human rights organization. After, you will be given a comment, where your task is to indicate whether a comment is \textit{toxic} or \textit{non-toxic}. If a comment is deemed \textit{toxic}, then you be provided with 5 additional labels (\textit{severe toxicity}, \textit{obscene}, \textit{threat}, \textit{insult} and \textit{identity attack}) to correctly identify and determine if a comment qualifies to be classified under one or more of the 5 additional labels.\\

\noindent \textbf{Human Annotator Protocol}
\begin{enumerate}
    \item Are you a member of the LGBTQIA+ community?
    
    \item If you responded ``no'' above, are you an LGBTQIA+ activist or ally?
    
    \item If you responded ``no'' above, please stop here.
    
    \item If you responded ``yes'' any of above question, given the extensive acronym list what is your identity, sexuality, and relationship? (Optional. This information is collected, but not saved, only for demographic purposes.)
    
    \item Are you willing to annotate several Reddit comments that contain \textit{stereotypes}, \textit{profanity}, \textit{vulgarity} and other harmful language geared towards LGBTQIA+ individuals?
    
    \item If you responded ``yes'' above, we must mention that if you believe you may become triggered or disturbed and cannot continue, please stop here.
    
    \item If you responded ``no'' above, please stop here.
    
\end{enumerate}

\noindent \textbf{Rating/ Sensitivity Protocol}
\begin{enumerate}
    \item As you responded ``yes'' a previous question,
    \begin{quote}\small{\textit{... Are you willing to annotate several Reddit comments that contain \textit{stereotypes}, \textit{profanity}, \textit{vulgarity} and other harmful language geared towards LGBTQIA+ individuals?}}\end{quote}
    You will be provided with 1000 comments which we have sampled from our \textit{binary} classification, and 5 additional labels. 
    
    \item For each comment, you will be tasked is to indicate whether a comment is \textit{toxic} or \textit{non-toxic}.\\ Is this comment toxic?
    \begin{enumerate}
        \item If you responded ``yes'' above, please select one or more of the appropriate labels provided considering these two classes, ``harmful : 1'' and ``non-harmful : 0''.  
        
        \item If you responded ``no'' above, please discard this comment. 
    \end{enumerate}
    
    \item Have you ever seen, heard, used or been called any of these Anti-LGBTQIA+ terms in a particular comment, for example, on social media or in-person? 
    
    \item If you responded ``yes'' above, do you feel triggered, disturbed or distressed reading this comment. (Optional. This information is collected, but is not saved, only for demographic purposes.)
    
\end{enumerate}
\noindent We would like to remind you that the objective of this study is not to cause more harm, but to create a safe and inclusive place that welcomes, supports, and values all LGBTQIA+ individuals both online and offline. However, due to the overall purpose of this study, we focus on online inclusivity.

\section{Data Breakdown}\label{sec:breakdown}

In this section, we display a breakdown of the data as the $\mathsf{toxicity}$ label is not an across-the-board label, but there exists a large amount of overlap between labels.\\

\noindent In total, there are 7459 toxicity comments. (75.12\% of all data.) \\
- 185 or 2.48\% were also severe toxicity. \\
- 1590 or 21.32\% were also obscene. \\
- 28 or 0.38\% were also threat. \\
- 2244 or 30.08\% were also insult. \\
- 4494 or 60.25\% were also identity attack. \\

\noindent In total, there are 185 severe toxicity comments. (1.86\% of all data.) \\
- 185 or 100.00\% were also toxicity. \\
- 185 or 100.00\% were also obscene. \\
- 13 or 7.03\% were also threat. \\
- 185 or 100.00\% were also insult. \\
- 184 or 99.46\% were also identity attack. \\

\noindent In total, there are 1590 obscene comments. (16.01\% of all data.) \\
- 1590 or 100.00\% were also toxicity. \\
- 185 or 11.64\% were also severe toxicity. \\
- 23 or 1.45\% were also threat. \\
- 1512 or 95.09\% were also insult. \\
- 1443 or 90.75\% were also identity attack. \\

\noindent In total, there are 28 threat comments. (0.28\% of all data.) \\
- 28 or 100.00\% were also toxicity. \\
- 13 or 46.43\% were also severe toxicity. \\
- 23 or 82.14\% were also obscene. \\
- 25 or 89.29\% were also insult. \\
- 27 or 96.43\% were also identity attack. \\

\noindent In total, there are 2244 insult comments. (22.60\% of all data.) \\
- 2244 or 100.00\% were also toxicity. \\
- 185 or 8.24\% were also severe toxicity. \\
- 1512 or 67.38\% were also obscene. \\
- 25 or 1.11\% were also threat. \\
- 2141 or 95.41\% were also identity attack. \\

\noindent In total, there are 4494 identity attack comments. (45.26\% of all data.) \\
- 4494 or 100.00\% were also toxicity. \\
- 184 or 4.09\% were also severe toxicity. \\
- 1443 or 32.11\% were also obscene. \\
- 27 or 0.60\% were also threat. \\
- 2141 or 47.64\% were also insult. \\

\section{Feature Distribution Plots}\label{sec:fd}

In this section, we display feature distribution \textit{i.e.}, visualization of the variation in the data distribution of each label. These distribution plots  represent the overall distribution of the continuous data variables.

\begin{figure}[!htbp]
    \centering
    \includegraphics[scale=0.6]{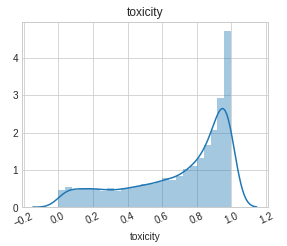}
    \caption{Toxicity feature distribution.}
    \label{fig:toxic}
\end{figure} 

\begin{figure}[!htbp]
    \centering
    \includegraphics[scale=0.6]{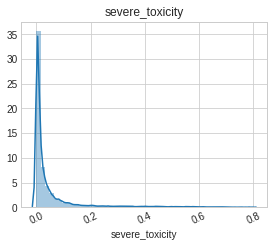}
    \caption{Severe Toxicity feature distribution.}
    \label{fig:severe}
\end{figure} 

\begin{figure}[!htbp]
    \centering
    \includegraphics[scale=0.6]{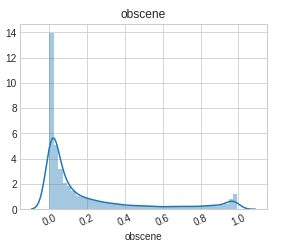}
    \caption{Obscene feature distribution.}
    \label{fig:obscene}
\end{figure} 

\begin{figure}[!htbp]
    \centering
    \includegraphics[scale=0.6]{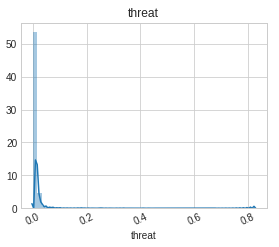}
    \caption{Threat feature distribution.}
    \label{fig:threat}
\end{figure} 

\begin{figure}[!htbp]
    \centering
    \includegraphics[scale=0.6]{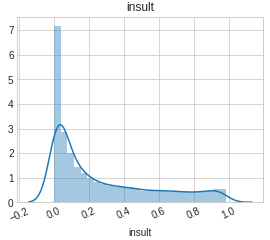}
    \caption{Insult feature distribution.}
    \label{fig:insult}
\end{figure} 

\begin{figure}[!htbp]
    \centering
    \includegraphics[scale=0.6]{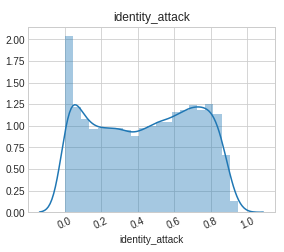}
    \caption{Identity attack feature distribution.}
    \label{fig:identity}
\end{figure}

\section{Word Contribution}\label{sec:word}

\textit{\textbf{\color{red}{Disclaimer:}}} Due to the overall purpose of the study, several terms in the figures may be offensive or disturbing (e.g. profane, vulgar, or homophobic slurs). These terms are not filtered as they are representative of essential aspects in the dataset.

\begin{figure*}[!htbp]
    \centering
    \includegraphics[scale=0.5]{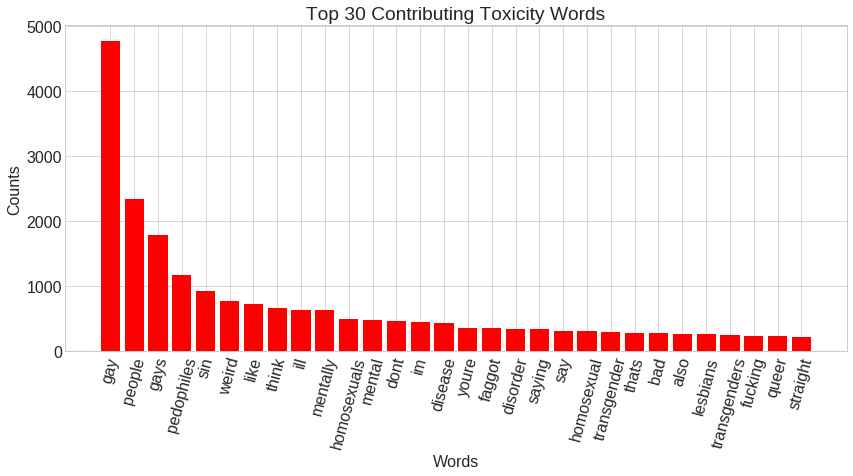}
    \caption{Top 30 most frequent words contributing to the Toxicity label.}
    \label{fig:toxic_words}
\end{figure*}

\begin{figure*}[!htbp]
    \centering
    \includegraphics[scale=0.5]{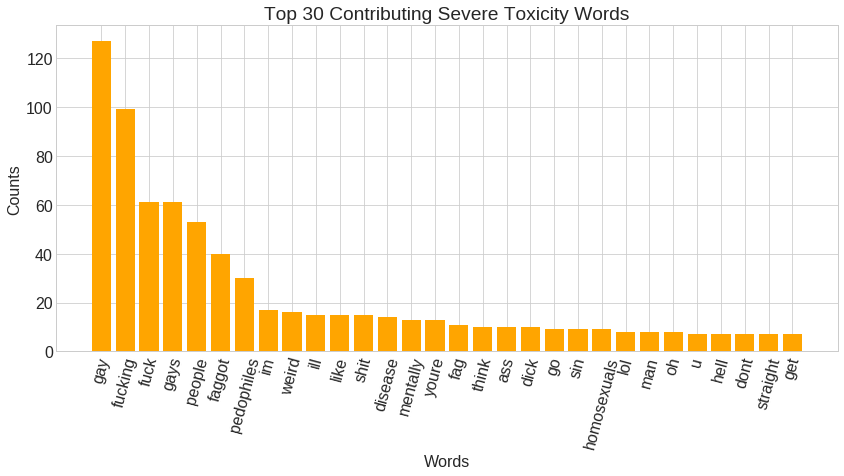}
    \caption{Top 30 most frequent words contributing to the Severe Toxicity label.}
    \label{fig:severe_words}
\end{figure*}

\begin{figure*}[!ht]
    \centering
    \includegraphics[scale=0.5]{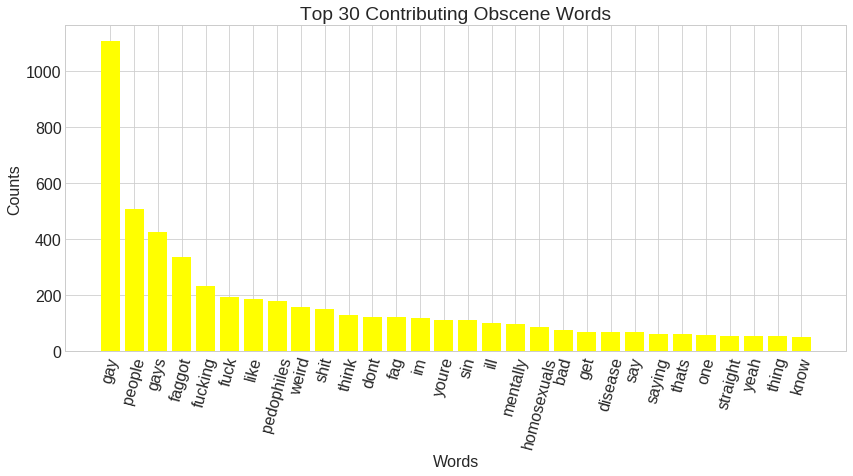}
    \caption{Top 30 most frequent words contributing to the Obscene label.}
    \label{fig:obscene_words}
\end{figure*}

\begin{figure*}[!ht]
    \centering
    \includegraphics[scale=0.5]{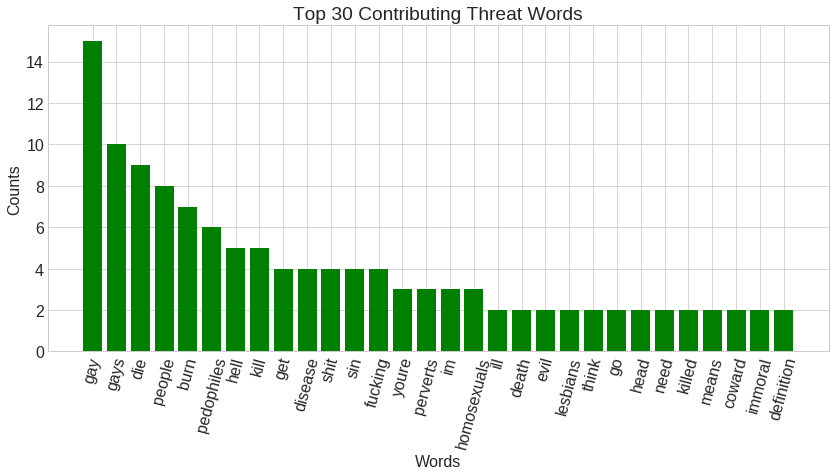}
    \caption{Top 30 most frequent words contributing to the Threat label.}
    \label{fig:threat_words}
\end{figure*}

\begin{figure*}[!ht]
    \centering
    \includegraphics[scale=0.5]{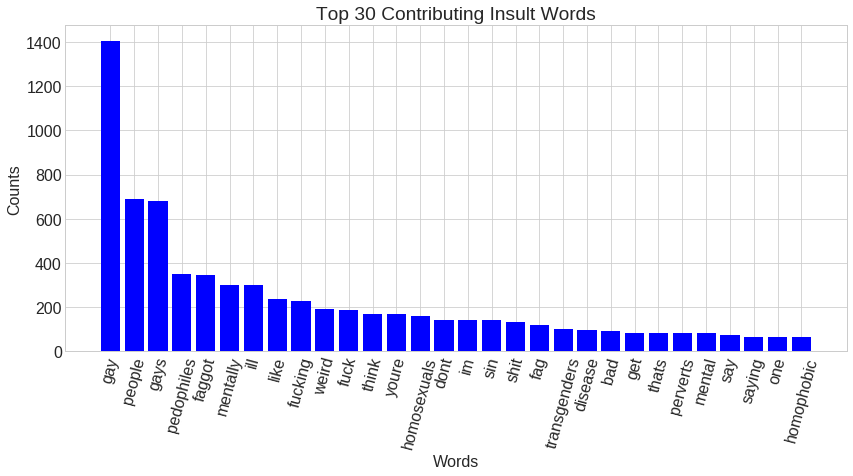}
    \caption{Top 30 most frequent words contributing to the Insult label.}
    \label{fig:insult_words}
\end{figure*}

\begin{figure*}[!ht]
    \centering
    \includegraphics[scale=0.5]{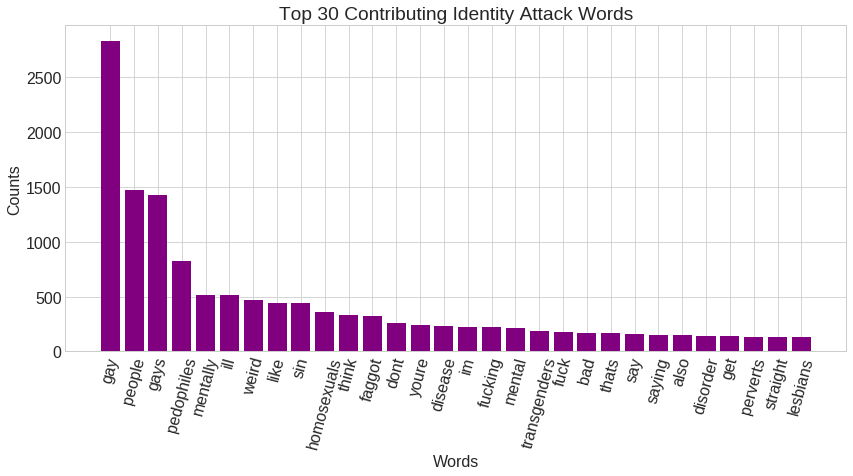}
    \caption{Top 30 most frequent words contributing to the Identity Attack label.}
    \label{fig:identity_words}
\end{figure*}

In this section, we demonstrate which words constitutes towards a “harmful” or “non-harmful” comment. In Figures~\ref{fig:toxic_words} -- \ref{fig:identity_words}, we display the top 30 most frequent words per label.

\end{document}